\pdfoutput=1

\documentclass[11pt]{article}

\usepackage{emnlp2021}

\usepackage{times}
\usepackage{latexsym}
\usepackage{hyperref}

\usepackage[T1]{fontenc}

\usepackage[utf8]{inputenc}
\usepackage{xcolor}
\usepackage{microtype}

\usepackage{graphicx}

\newcommand\Mark[1]{\textsuperscript{#1}}

\title{Does language help generalization in vision models?}

 \author{
 Benjamin Devillers\Mark{1}, Bhavin Choksi\Mark{2}, Romain Bielawski\Mark{1}, Rufin VanRullen\Mark{1,2} \\
    \begin{tabular}{c}
        \\
        \Mark{1} Artificial and Natural Intelligence Toulouse Institute, Université de Toulouse, France \\
        \texttt{\{firstname.lastname\}@univ-tlse3.fr} \\
        \\
        \Mark{2} CerCO, CNRS UMR5549, Toulouse \\
        \texttt{\{firstname.lastname\}@cnrs.fr}
    \end{tabular}
}

\begin{document}
\maketitle

\begin{abstract}
Vision models trained on multimodal datasets can benefit from the wide availability of large image-caption datasets. A recent model (CLIP) was found to generalize well in zero-shot and transfer learning settings. This could imply that linguistic or ``semantic grounding'' confers additional generalization abilities to the visual feature space. Here, we systematically evaluate various multimodal architectures and vision-only models in terms of unsupervised clustering, few-shot learning, transfer learning and adversarial robustness. In each setting, multimodal training produced no additional generalization capability compared to standard supervised visual training. We conclude that work is still required for semantic grounding to help improve vision models.
\end{abstract}

\section{Introduction}

Learning vision models using language supervision has gained popularity \cite{ quattoni2007learning, srivastava2012multimodal, frome2013devise, joulin2016learning, pham2019found, desai2021virtex, hu2021transformer, radford2021learning,sariyildiz2020icmlm} for two main reasons: firstly, vision-language training allows to build massive training datasets from readily available online data, without manual annotation; 
secondly, language provides additional semantic information that cannot be inferred from vision-only datasets, and this could help with semantic grounding of visual features.

Recently~\cite{radford2021learning} introduced CLIP, a language and vision model that shows outstanding zero-shot learning capabilities on many tasks, and compelling transfer-learning abilities. A recent report \cite{goh2021multimodal} showed that CLIP produces neural selectivity patterns comparable to ``multimodal'' concept cells observed in the human brain \cite{quiroga2005invariant, reddy2014concept}. 
From these results, it is tempting to assume that CLIP's generalization properties stem from semantic grounding provided by the joint vision-language training. 
 


Here, we show that CLIP and other vision-language models do not perform better than vision-only, fully supervised models on a number of generalization settings and datasets. Representation similarity~\cite{10.3389/neuro.06.004.2008} analysis reveals that the multimodal representations that emerge through vision-language training are different from \emph{both} linguistic and visual representations--and thus possibly unsuitable for transfer-learning to new visual tasks. In conclusion, additional work on linguistic grounding is still needed, if it is to improve generalization capabilities of vision models.

We provide our code for reproducibility\footnote{\url{https://github.com/bdvllrs/generalization-vision}}.

\begin{figure}[t]
    \centering
    \includegraphics[width=\linewidth]{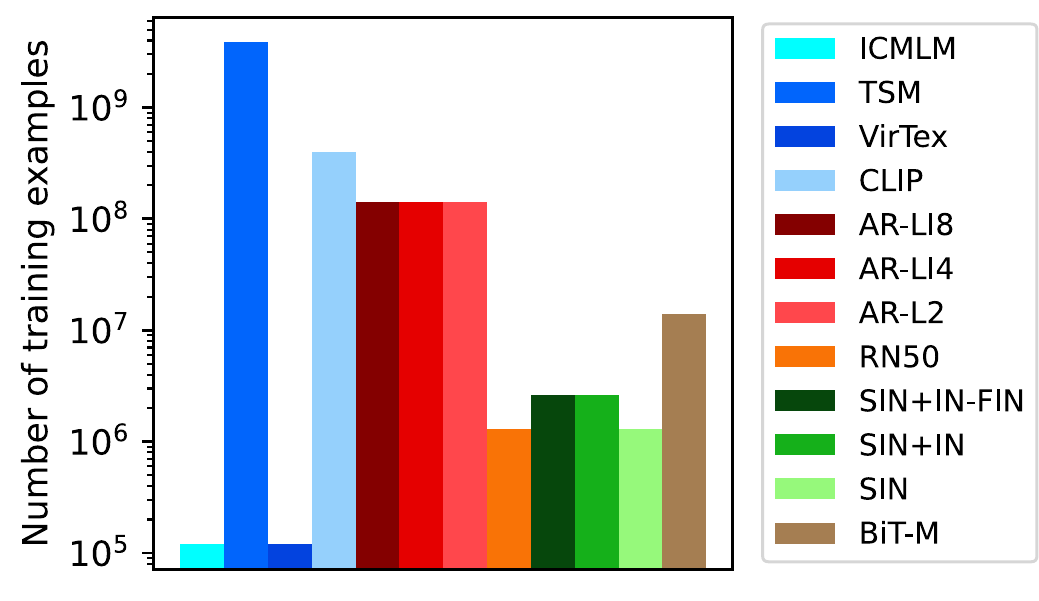}
    \caption{Size of the training dataset used by the models (number of images, in log scale). ICMLM and VirTex are trained on CoCo, TSM on HowTo100M, CLIP on a (not publicly available) scrape of the internet, RN50 is trained on ImageNet-1k, the AR models and SIN models are trained on augmented versions of ImageNet-1k and BiT-M is trained on ImageNet-21k.}
    \label{fig:dataset-size-training}
\end{figure}

\section{Models}

We use a number of publicly available vision, text or multimodal pretrained models, and compare their representations and generalization abilities. To facilitate interpretation and comparisons between the models, Figure~\ref{fig:dataset-size-training} reports the training dataset size for each of the visual models (including the vision-language models). They are all based on the same backbone (a ResNet50 architecture). 

In CLIP, the authors train the joint embedding space of a visual network (hereafter called simply CLIP) and a language network (hereafter called CLIP-T) using contrastive learning on 400M image-caption pairs. Note that in the present paper, the visual backbone of CLIP is a ResNet50, even though the visual-transformer-based CLIP model could reach higher performance; this choice allows for a fair comparison with the other visual models that are all based on the ResNet50 architecture. In addition, we also consider TSM~\cite{alayrac2020self}, another multimodal network trained with a contrastive loss on video, audio and text inputs from the HowTo100M dataset \cite{miech2019howto100m} (containing more than 136M video clips with captions. For training, the authors effectively used 120M video clips of 3.2s sampled at 10 fps). The effects of CLIP's and TSM's contrastive training paradigm can be compared with VirTex and ICMLM---two other recent multimodal networks. In VirTex, the visual feature representations are optimized for an image captioning task~\cite{desai2021virtex}, and for a text-unmasking task in ICMLM~\cite{sariyildiz2020icmlm}. Such text-based objectives aim to provide a form of linguistic grounding using significantly fewer images than CLIP (VirTex and ICMLM models are trained on the COCO dataset \cite{Lin_2014} with approximately 120K captioned images).

To understand the potential effects of linguistic training, we compare the multimodal networks to vision-only networks. We include a baseline architecture (ResNet50) trained on ImageNet-1K~\cite{he2016deep} (1.3M labelled images). Second, we consider a similar architecture (ResNet50 backbone) called BiT-M~\cite{kolesnikov2019big}, trained on ImageNet-21K, a much larger dataset (14M labelled images). 

While generalization and robustness properties can often be derived from access to large labelled image datasets (as in BiT-M), obtaining such labels is costly. An alternative is to train models with additional datapoints based on assumptions about the real-life data distribution--as done, e.g., with adversarial training. In this study, we use the Adversarially Robust (AR) ResNet50 models  provided by~\cite{engstrom2019adversarial}, trained on the 1.3M ImageNet training set plus 110 adversarial attacks of each image (i.e. more than 140M images overall). The different model variants (AR-L2, AR-LI4, AR-LI8) correspond to distinct adversarial attacks (refer to~\cite{engstrom2019adversarial} for more details).
This adversarial training was found to produce more perceptually aligned features and to improve generalization (e.g. transfer learning) in some settings~\cite{salman2020adversarially}. Another such technique was used for StylizedImageNet (SIN) models~\cite{geirhos2018imagenettrained}, where a variant of the ImageNet dataset (1.3M images) was designed via style-transfer to specifically reduce the network's reliance on texture information. The authors provide weights for models that are (i) only pretrained for SIN images (SIN), (ii) trained on SIN and ImageNet (SIN+IN) combined, or where (iii) a SIN+IN model is finetuned on ImageNet (SIN+IN-FIN).

For the vanilla ResNet50, SIN, AR and BiT-M models, we use activations after the final average pooling operation as feature representations. 
Although all these models share a ResNet50 backbone, there are minor differences in their implementations. 
We assume that such small architectural differences would not dramatically affect the feature spaces learned by these models.

Finally, we also use two text-only language models, GPT-2~\cite{radford2019language} and BERT~\cite{devlin2018bert}, in our feature-space comparisons. As these models are not designed to process visual inputs, they cannot be tested on visual generalization; but we can use their representations of class \textit{labels} (or sentence captions) as a basis for comparison with visual or multimodal network representations. In a similar way, the language stream of the CLIP model (CLIP-T) can be treated as a third language model for our comparisons.

\begin{figure*}[t]
    \centering
    \includegraphics[width=\linewidth]{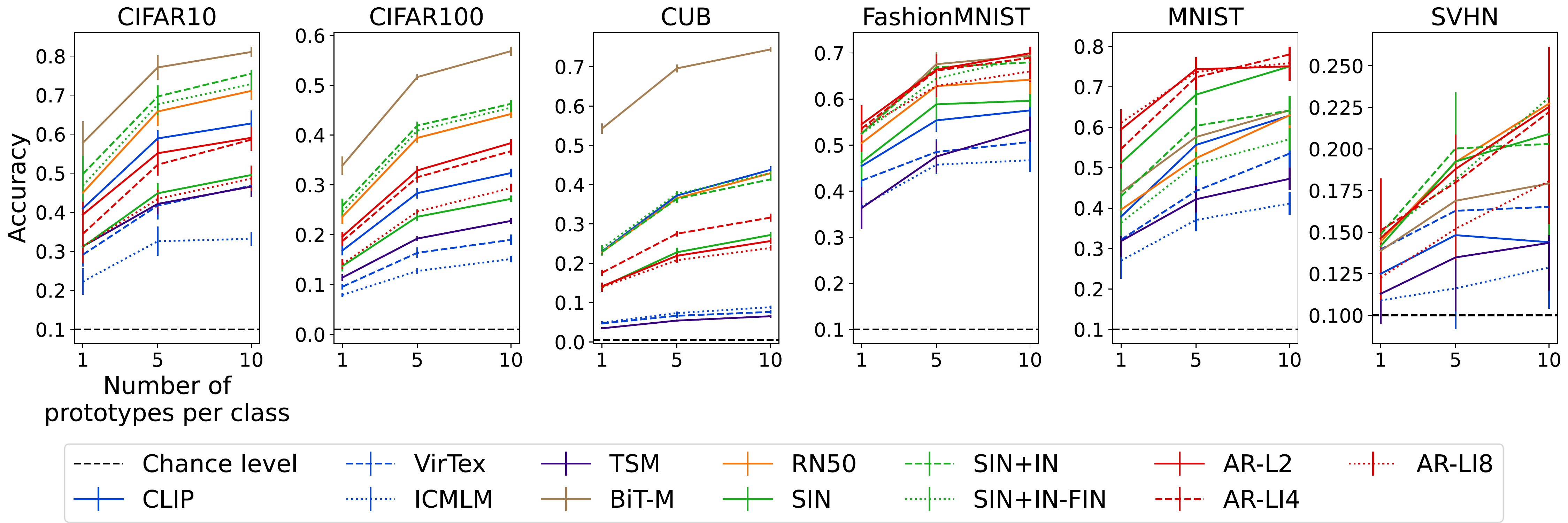}
    \caption{1-shot, 5-shot and 10-shot accuracy over our evaluation datasets. Multimodal networks (ICMLM, VirTex, CLIP, TSM, in blue) have typically worse performance than the other models for all datasets.}
    \label{fig:few-shot}
\end{figure*}
\begin{figure*}[t]
    \centering
    \includegraphics[width=\linewidth]{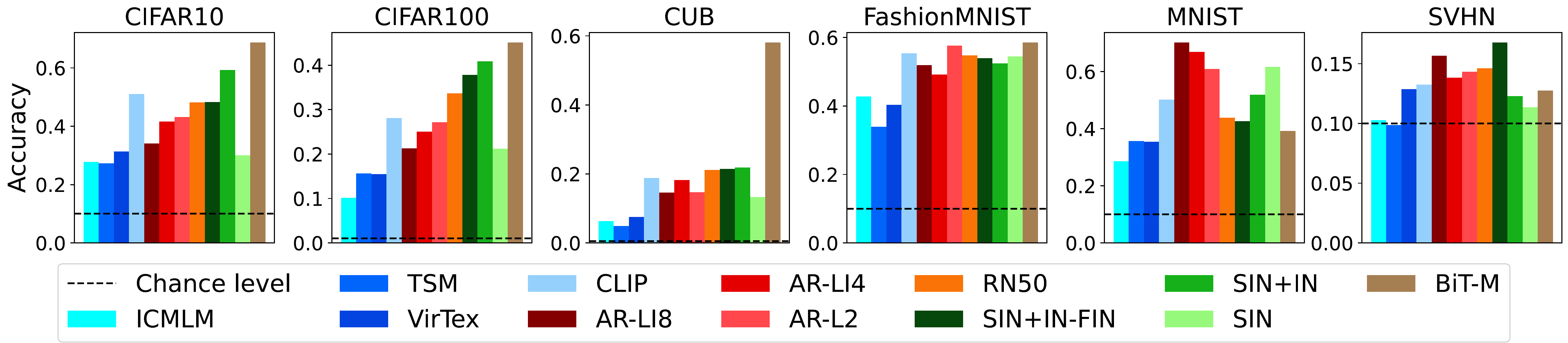}
    \caption{Unsupervised clustering accuracy over our evaluation datasets. Clustering is obtained using Scikit-learn Spectral Clustering algorithm. Multimodal networks (ICMLM, VirTex, CLIP, TSM, in blue) are worse than vision-only models (in various colors) on average.}
    \label{fig:clustering}
\end{figure*}

\section{Generalization tasks}
In \cite{radford2021learning}, CLIP was systematically tested in a zero-shot setting. However, this requires a language stream to describe the different possible targets, which is not available in standard vision models.
To compare the generalization capabilities of multimodal and vision-only models, we thus focus on few-shot, transfer and unsupervised learning. In each case, we evaluate performance on MNIST \cite{lecun1998gradient}, CIFAR10, CIFAR100 \cite{krizhevsky2009learning}, Fashion-MNIST \cite{xiao2017/online}, CUB-200-2011 (CUB) \cite{WahCUB_200_2011} and SVHN \cite{netzer2011reading}\footnote{For more details on these datasets, see appendix \ref{appendix:datasets}.}. These datasets test generalization capabilities for natural images of various classes.



\begin{figure*}[t]
    \centering
    \includegraphics[width=\linewidth]{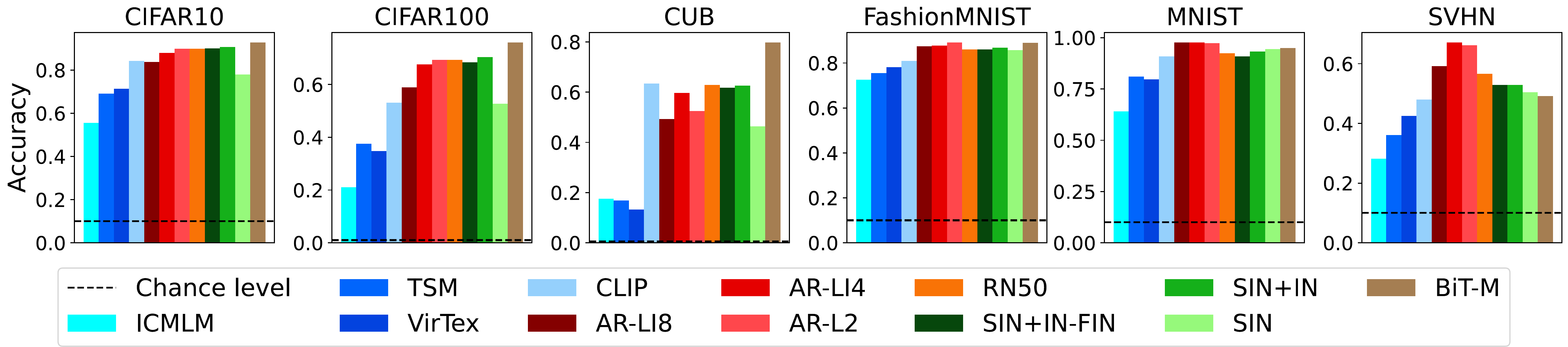}
    \caption{Transfer learning accuracy over our evaluation datasets. For each dataset and model, we train a linear layer to classify the models' visual features. Multimodal networks (ICMLM, VirTex, CLIP, TSM, in blue) have worse performance accuracy than vision-only models (in various colors).}
    \label{fig:transfer-learning}
\end{figure*}

\begin{figure*}[t]
    \centering
    \includegraphics[width=\linewidth]{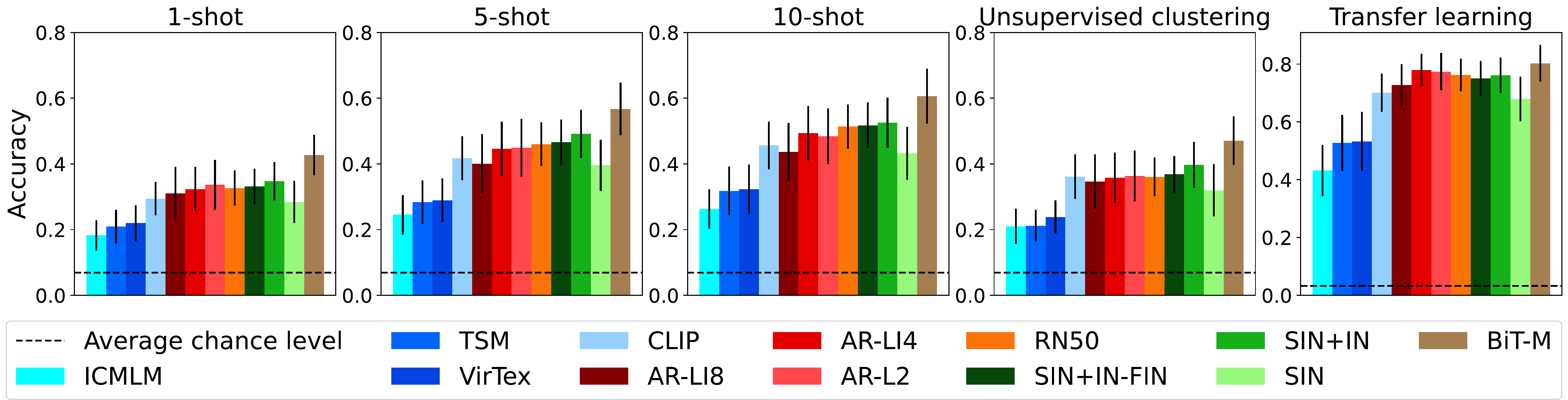}
    \caption{Average performance of the models across datasets, with standard error of the mean, for the various generalization tasks (few-shot learning, unsupervised clustering, transfer learning). 
    Multimodal networks (ICMLM, VirTex, CLIP, TSM in blue) have worse generalization accuracy across all tasks.
    }
    \label{fig:averaged-performance}
\end{figure*}

\subsection{Few-shot learning}
\label{section:few-shot}
As a first generalization experiment, we compare few-shot learning accuracy. For this experiment, we directly pass $N$ randomly selected samples for each class ($N$-shot learning) through the pretrained models to obtain a feature representation for each sample. Then, we define a class prototype by averaging the feature representations of all the samples in each class. We measure the performance of vision-only and text-vision models for $N=$ 1, 5 and 10. Each time, the reported performance is averaged over 10 trials with different class prototypes (i.e., different random selection of samples). 
Figure~\ref{fig:few-shot} shows the performance of each model on each dataset. For CIFAR10, CIFAR100 and CUB (all the natural images datasets), BiT-M has the best accuracy. On the other hand, ICMLM, VirTex, CLIP and TSM do not perform better than the vision-only models.

Figure~\ref{fig:averaged-performance} shows the average performance of each model across datasets, in the leftmost 3 panels.

\subsection{Unsupervised clustering}
Our second generalization test is an unsupervised clustering task over the same datasets.
For this, we apply an out-of-the-box spectral clustering algorithm ~\cite{scikit-learn} using the cosine of two feature vectors as a metric. We provide the number of required clusters (number of classes) to the clustering algorithm: this ensures that all classes are represented by a cluster.
The clusters are computed only on the test-sets.

 To compute the accuracy on the prediction, we need to assign labels to each cluster. To do so, we use a greedy algorithm:
we first choose the cluster containing the most elements in common with a given class and assign it the corresponding label. We then continue with the second cluster that has the most elements in common with another class, and so on until all clusters have been labelled. 

Figure \ref{fig:clustering} shows the unsupervised clustering performance on individual datasets. It shows a similar ranking to the few-shot learning task where BiT has the best performance overall and the visio-linguistic models lag behind the vision-only models.
Figure \ref{fig:averaged-performance} panel 4 (from left) shows the performance of the unsupervised clustering algorithm averaged over all datasets.

\subsection{Transfer learning}
To further evaluate the models' generalization properties, we use a transfer learning setting as described in \cite{salman2020adversarially}. We use the same datasets as in the other tasks, each time training a linear probe using the Adam optimizer. We train each linear probe for 20 epochs with a learning rate of 1e-3 and a weight decay of 5e-4.

Fig~\ref{fig:transfer-learning} shows the performance of the models on this task, separately for each dataset, and Fig~\ref{fig:averaged-performance} (rightmost panel) reports the average across datasets. Multimodal networks fail again to improve generalization.

\begin{figure}[h!]
    \centering
    \includegraphics[width=\linewidth]{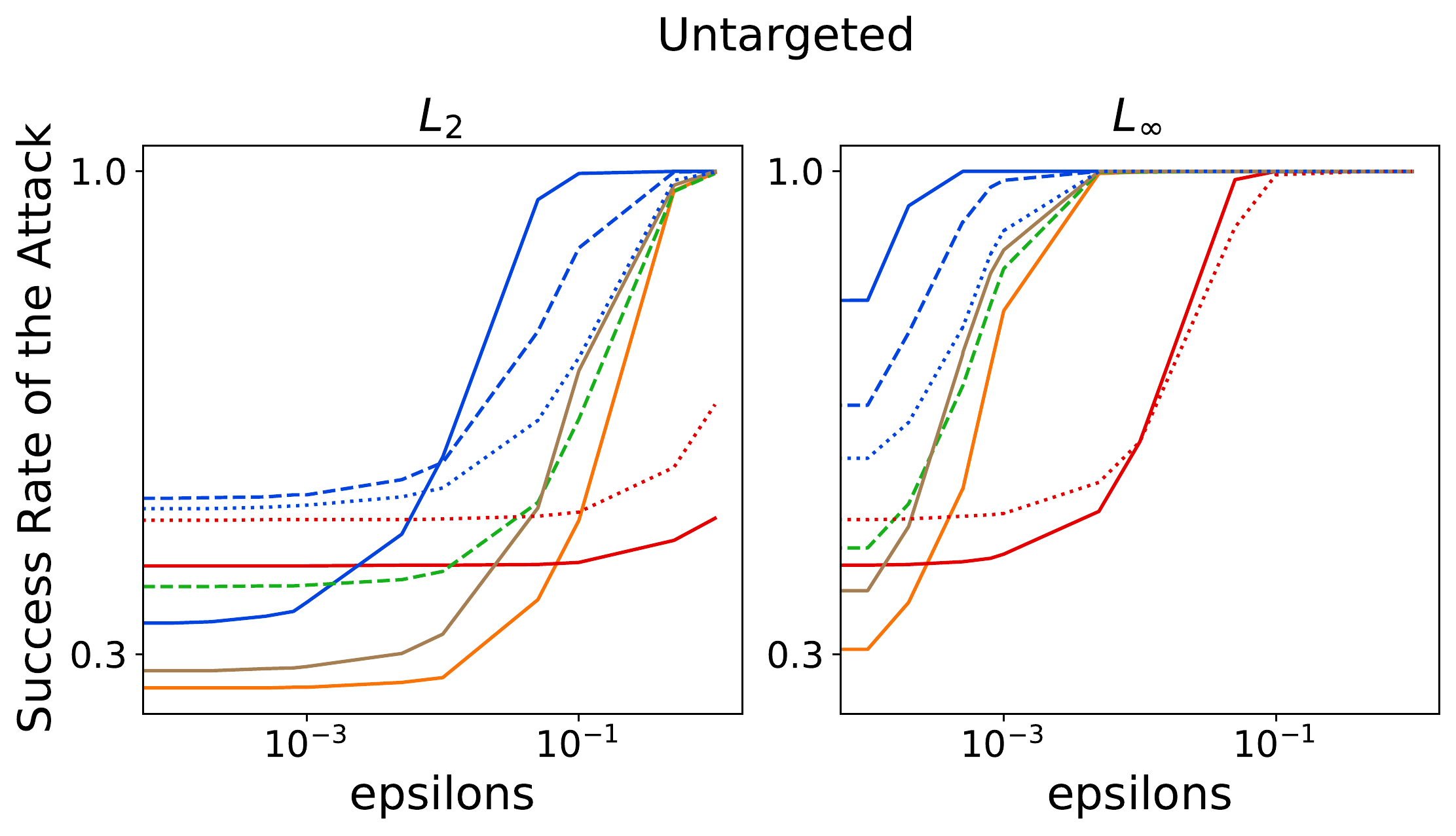}
    \includegraphics[width=\linewidth]{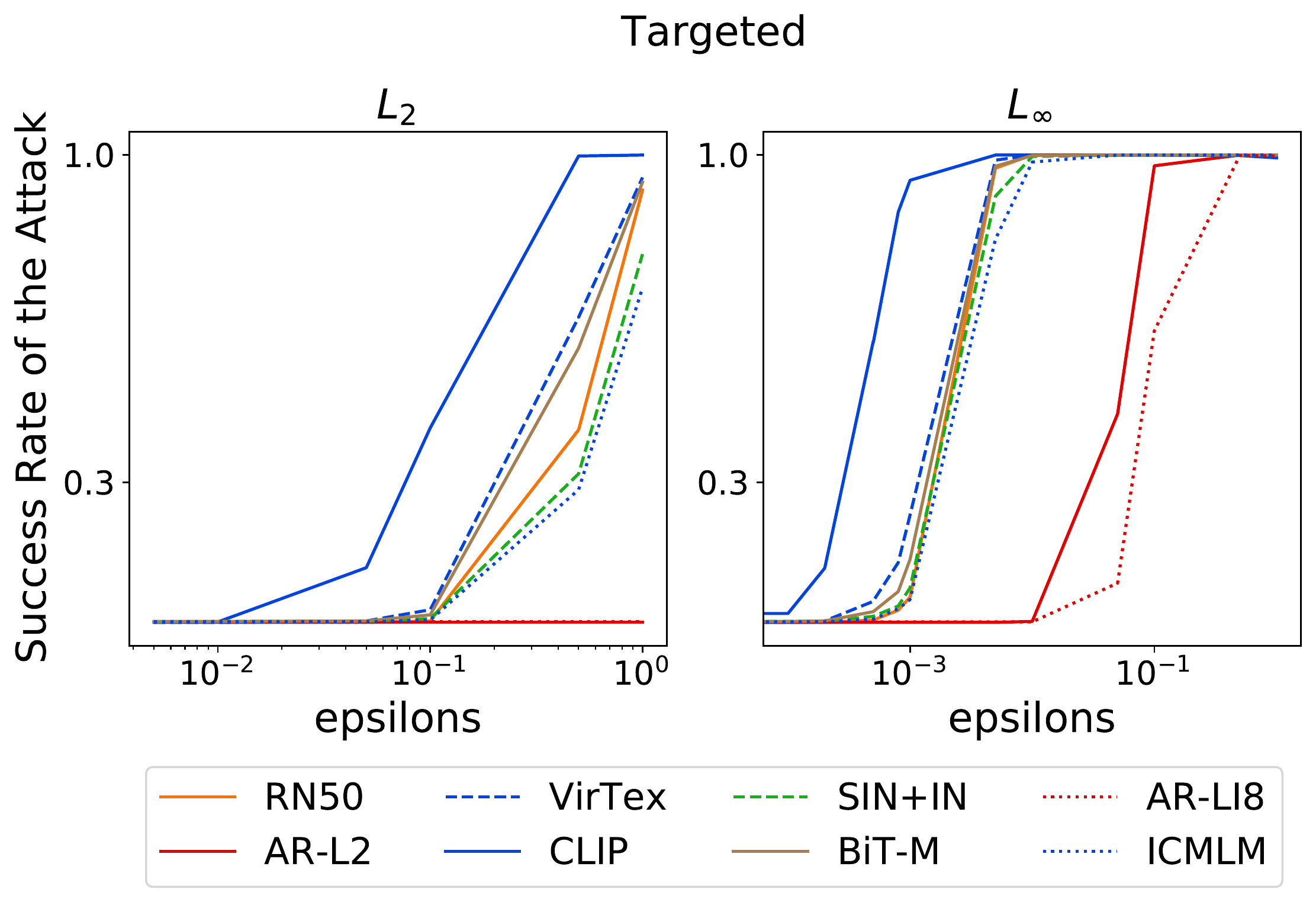}
    \caption{Robustness of some of the models to untargeted (top) and targeted (bottom) random projected gradient descent (RPGD) attacks for varying epsilons, with $L_2$ (left) or $L_{\infty}$ norm (right). AR models are robust by design. Multimodal networks (CLIP, VirTex) are less robust than vision-only models (RN50, SIN+IN, BiT-M).}
    \label{fig:robustness}
\end{figure}

\subsection{Robustness to adversarial attacks}

Another important test for generalization is the robustness to input perturbations (a form of out-of-distribution generalization). Here, we compare the adversarial robustness of different models against untargeted and targeted random projected gradient descent (RPGD) attacks~\cite{madry2017towards}. We use $L_{2}$ and $L_{\infty}$ norms to distinguish any norm-specific effects. Figure~\ref{fig:robustness} shows the success rate of the 100-step RPGD attacks on 1000 images taken from the ImageNet validation set. We use the foolbox API~\cite{rauber2017foolbox} to perform all the attacks with configurations provided by ~\cite{robustness}.


\subsection{Summary}
Overall, models trained with multimodal information (CLIP, VirTex, ICMLM, TSM) do not achieve better performance than the visual-only ResNet-based models. This systematic observation across multiple image datasets and generalization tasks (including few-shot, transfer and unsupervised learning, as well as adversarial robustness) goes against the assumption that linguistic grounding should help generalization in vision models.

Among the multimodal networks, CLIP does indeed appear to be more generalization-efficient than VirTex, ICMLM and TSM. As mentioned in \cite{radford2019language}, directly predicting highly variable text captions (as done in VirTex or ICMLM) is a difficult task that does not scale well. CLIP (and TSM) avoid generating text, relying instead on a contrastive loss between visual and linguistic embeddings. However, even with the potential benefits provided by this contrastive loss, CLIP (and TSM) do not outperform the vision models.

Finally, BiT-M, a simple vision-only model trained on a very large labelled dataset, turns out to be the overall best performing model for few-shot learning, unsupervised clustering and transfer learning, and on par with the standard ResNet50 for adversarial robustness.

 Although these results are fairly consistent across datasets, there are still some differences.
 
 For the CUB dataset, BiT-M largely outperformed the other models. This result is to be expected as the bird species in CUB are also part of ImageNet-21K labels. Then, among visio-linguistic models, CLIP is the only one competitive with the remaining visual models on this dataset.

 MNIST and SVHN require classification of digits. According to \cite{radford2019language}, CLIP should be able to generalize to this task, as its training set included numerous images with text and digits. 
 Indeed we observe that CLIP can perform as well as some of the vision models for these datasets. However, SIN and AR models perform generally better than other models.

 For datasets with more natural images (CIFAR, FashionMNIST, CUB), vision models are generally better than their visio-linguistic counterparts.

\section{Model comparison}
To better understand the similarities and differences between the feature spaces learned by the various models, we now compare them using Representational Similarity Analysis (RSA) \cite{10.3389/neuro.06.004.2008}.

\paragraph{Method} 
RSA is a comparison method originally used to compare fMRI data. It allows us to compare different models (with different latent space dimensions, norms, ...) which share the same structure.

\begin{figure}[t]
    \centering
    \includegraphics[width=\linewidth]{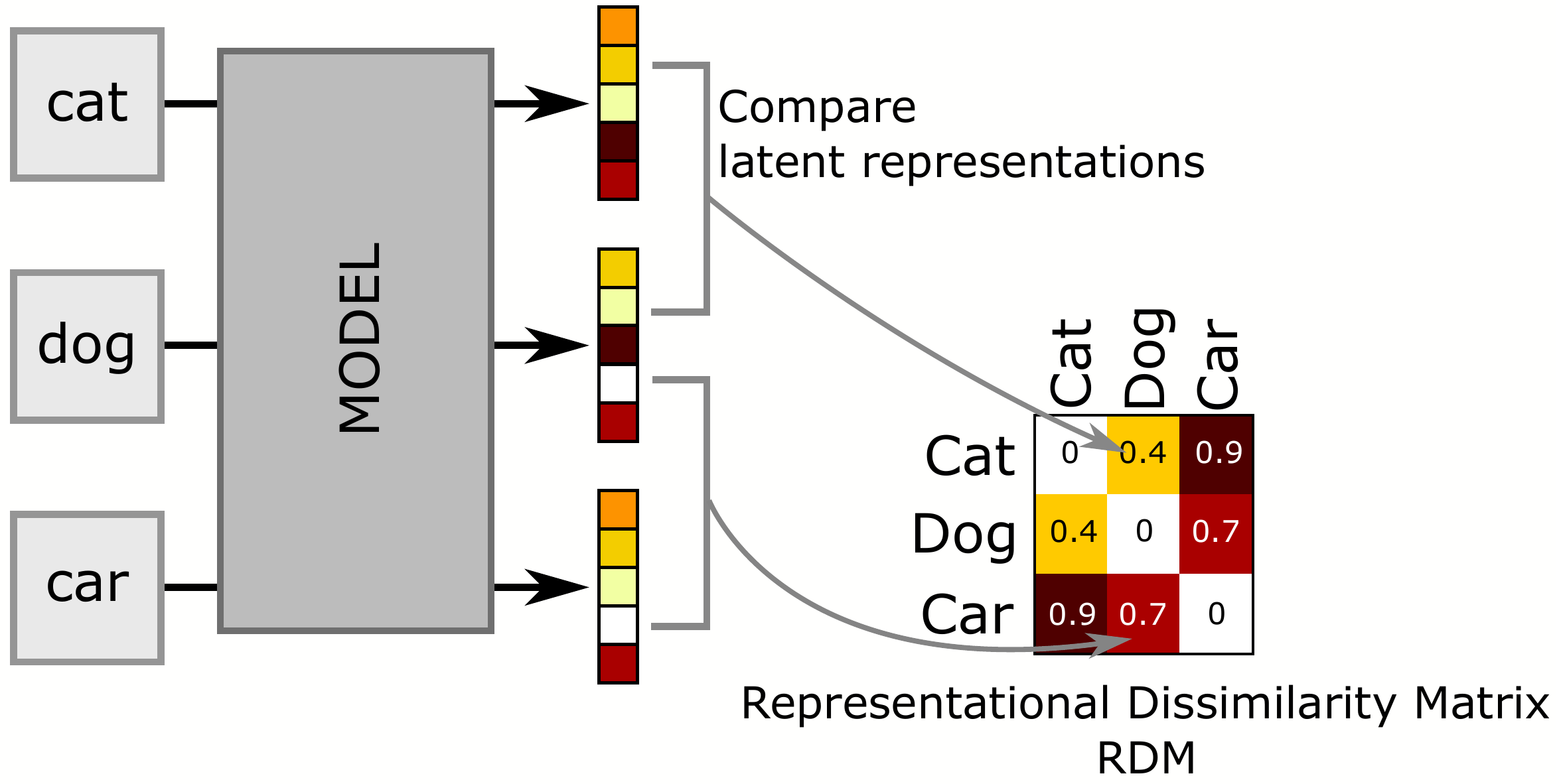}
    \caption{How to compute representational Dissimilarity Matrices (RDMs). RDMs are built from the model's embedding space. The RDMs can then be used for a Representational Similarity Analysis by comparing them using a Pearson Correlation.}
    \label{fig:rsa-diagram}
\end{figure}

This works by comparing the models' \textit{Representational Dissimilarity Matrix} (RDM).
RDMs are obtained by computing the 2 by 2 distances for each class of the latent representations (see figure~\ref{fig:rsa-diagram}).
More specifically, for each visual model, we define for each class the set $\mathcal{F}_c$ containing the feature vectors of all the images of class $c$, its average $\bar{f}_c$ and its standard deviation $\sigma_c$. The RDM matrix is then defined as $[RDM_{i,j}]$ where 

\begin{equation}
   RDM_{i,j} = \left\|\frac{\bar{f}_i - \bar{f}_j}{\sqrt{\frac{\sigma_i^2}{|\mathcal{F}_i|} + \frac{\sigma_j^2}{|\mathcal{F}_j|}}}\right\|_2 
\end{equation}

for each pair of class $(i,j)$.

We use the norm of the unequal variance t-test \cite{welch1947generalization} as our distance metric between the latent representations, because it allows to normalize the distances between class centroids with respect to their variances. Indeed, each class is represented by a cluster of latent vectors of different sizes.

In the case of language models (all transformer-based), we use as latent representations, the encoding of the sentence ``a photo of x.'' where we replace ``x'' by the corresponding label. 
We then use the contextualization of the label as the text feature vector. 
Compared to the vision models, there is  only one representation per class (only one sentence per class) hence a lack of variance associated with the feature vector of each class. As a result,  
the distance used in the RDM matrix becomes an $L_2$ norm.

\begin{figure*}[ht!]
    \centering
    \includegraphics[width=0.66\linewidth]{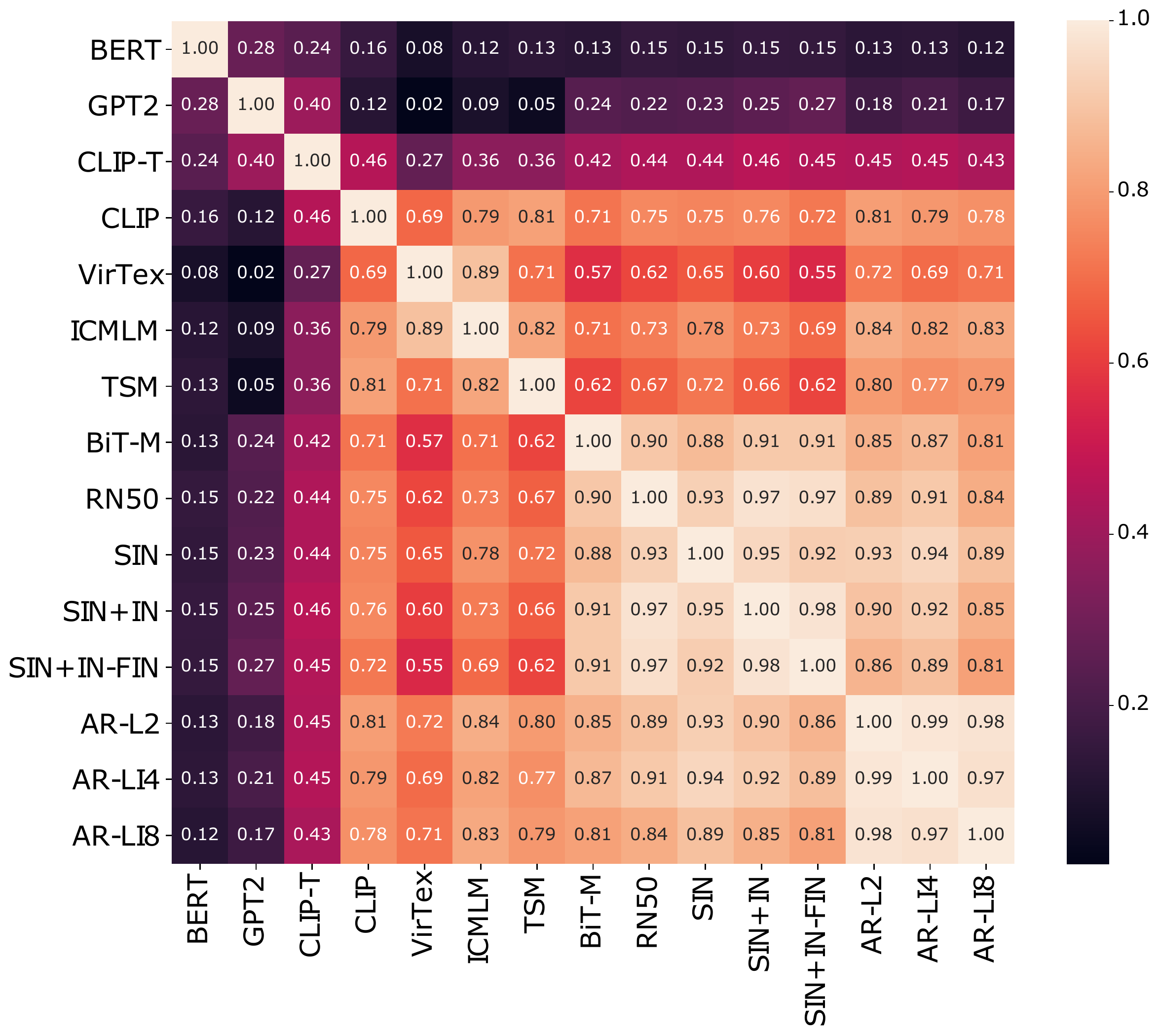}
    \caption{Correlations of the RDMs of our evaluation models. The RDMs are computed as explained in Fig~\ref{fig:rsa-diagram} using the ImageNet dataset.}
    \label{fig:rdm}
\end{figure*}

The RDM matrix obtained with this method contains the respective distances between pre-defined concepts (in our case the 1000 classes of ImageNet). 
RDMs can therefore be considered as a standardized representation of latent spaces.
This means that we can compare our models' representations by computing the Pearson correlation between their respective RDMs. The corresponding comparison matrix, for all pairs of models, is illustrated in Fig~\ref{fig:rdm}.

\begin{figure*}[ht!]
    \centering
    \includegraphics[width=0.66\linewidth]{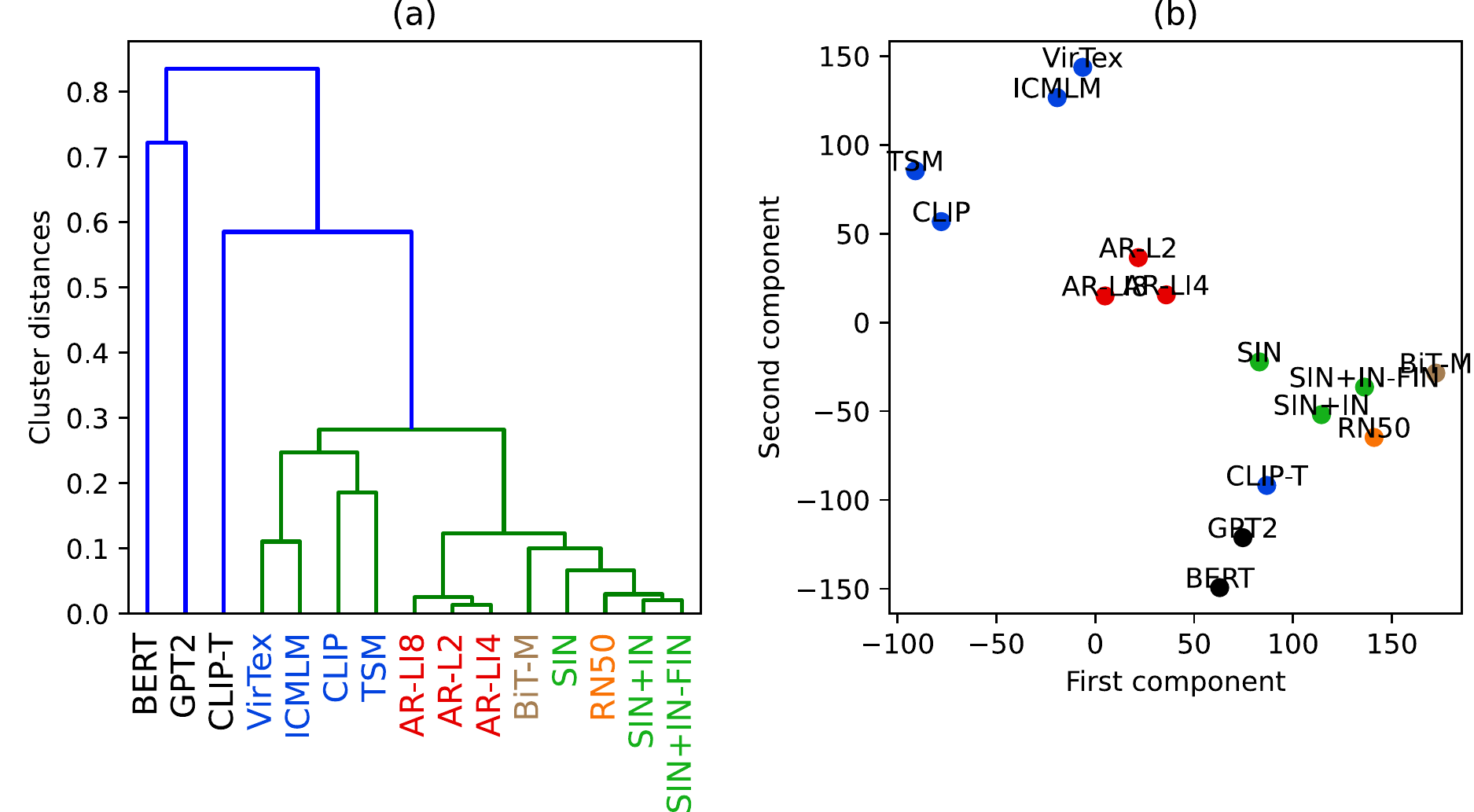}
    \caption{(a) Dendrogram of a hierarchical clustering of the RDMs. (b) t-SNE of the RDMs.}
    \label{fig:tsne-dendrogram}
\end{figure*}

\paragraph{Results}
Figure \ref{fig:tsne-dendrogram} shows the results of a hierarchical clustering (a) or t-SNE~\cite{van2008visualizing}  embedding (b) of the RDMs using Pearson correlation as a distance. 
Looking at the dendrogram, all the vision-only models are very close to one another with a maximum distance <0.2. Then, multimodal models stand a bit further (CLIP, TSM, VirTex, ICMLM); and finally, CLIP-T and the language models (BERT, GPT2) are the furthest away. This indicates that the language supervision (contrastive embedding, text-generation or text-unmasking objectives) has changed the structure of the ResNet latent space for CLIP, TSM, VirTex and ICMLM models (respectively). Yet these multimodal models are not truly linguistic either, as they are very distant also from the standard language models.

This conclusion is also supported by the t-SNE plot, showing a cluster of BiT-M, RN50 and SIN vision models, a second cluster with the AR models, and further along the same direction, the multimodal networks (CLIP, VirTex, ICMLM, TSM). Note that, although this arrangement might suggest that multimodal networks possess adversarial robustness properties in common with AR models, this suggestion was not supported by our tests using actual adversarial attacks (Fig \ref{fig:robustness}). Finally, the language models (BERT, GPT2 and CLIP-T) are separated from the rest, along a distinct direction. Overall, the analysis suggests that multimodal representations are neither visual nor linguistic, but surprisingly, \emph{not really in-between either}\footnote{Of course, we describe multimodal networks as \emph{neither visual nor linguistic}, but this is to be understood in relative terms--they are \emph{relatively} far from both visual models and linguistic models. In absolute terms, there is always a reasonable amount of similarity between multimodal networks and certain visual or linguistic models.}. 
 This is surprising as we should expect that representations trained with access to both vision and language would derive information from both modalities, and consequently end up somewhere in-between purely visual and purely textual representations.

\section{Performance on linguistic tasks}
This suggestion might be further supported by evaluating the usefulness of the learned visual representations on \emph{linguistic} tasks. According to the above findings, visual representations obtained via multimodal training may fare no better than vision-only representations. To test this, for each vision model, we collect the ImageNet features for each image class, and train a standard word embedding (Skip-Gram method) while constraining the class label words to these visual feature vectors. The resulting linguistic space will thus capture some of the structure of the vision model's latent space.

\subsection{Method}

\paragraph{Architecture}
We train Skip-Gram models \cite{mikolov2013efficient} on Wikipedia using the Gensim library \cite{rehurek_lrec}. Before training, some of the embedding vectors (corresponding to the ImageNet class labels) are set to the latent representations of a vision model, and frozen during training.
This training procedure forces the word embedding space to adopt a similar structure to the vision model's latent space (at least for the frozen words, i.e. the class labels).

\paragraph{Visual words}
We denote `visual word embeddings' (resp. visual words) as the word embeddings (equivalent to the visual feature vectors) obtained from the vision models (resp. the associated word token) on ImageNet classes.
Some of the classes are composed of multiple words (e.g. ``great white shark''). We leverage the WordNet \cite{miller1998wordnet} structure of ImageNet classes to only keep the hypernym of the class that contains only one word (e.g. ``great white shark'' becomes ``shark''). All of the ImageNet categories that have the same one-word hypernym are grouped together into one unique hyperclass. For instance, the ``shark'' hyperclass contains the classes ``great white shark'' and ``tiger shark''.
Finally, to obtain the visual word embeddings, we average the visual representation of all the images of each hyperclass from the ImageNet validation set. This gives a total of 824 visual words.

Besides, we choose a vocabulary of 20,000 words (taken from the most frequent tokens in Wikipedia). Only 368 visual words are among the 20,000 most frequent words, so we extend our vocabulary to also contain the 456 other visual words, resulting in a total vocabulary of 20,456 words.

\paragraph{Embedding dimension}
Since the vision models do not all share the same feature dimensions, in order to compare all Skip-Gram models, we reduce the dimensionality of the feature spaces of all vision models to 300 dimensions using a PCA. The PCA is computed using the visual features of all images in the ImageNet validation set. Consequently, the Skip-Gram word embeddings are trained with 300 dimensions. 

\paragraph{Training}
We train the Skip-Gram models for 5 epochs, using the standard negative sampling strategy.
We use window sizes of 5 words and a learning rate of 1e-3. We use the ``\texttt{vectors\_lockf}'' feature of the Gensim library to freeze certain word embeddings during training.

For the dataset, we use a recent dump of Wikipedia and we split it into two sets containing 80\% and 20\% of the articles for the training and validation sets.


\subsection{Evaluation}
\begin{figure*}[t!]
    \centering
    \includegraphics[width=0.64\linewidth]{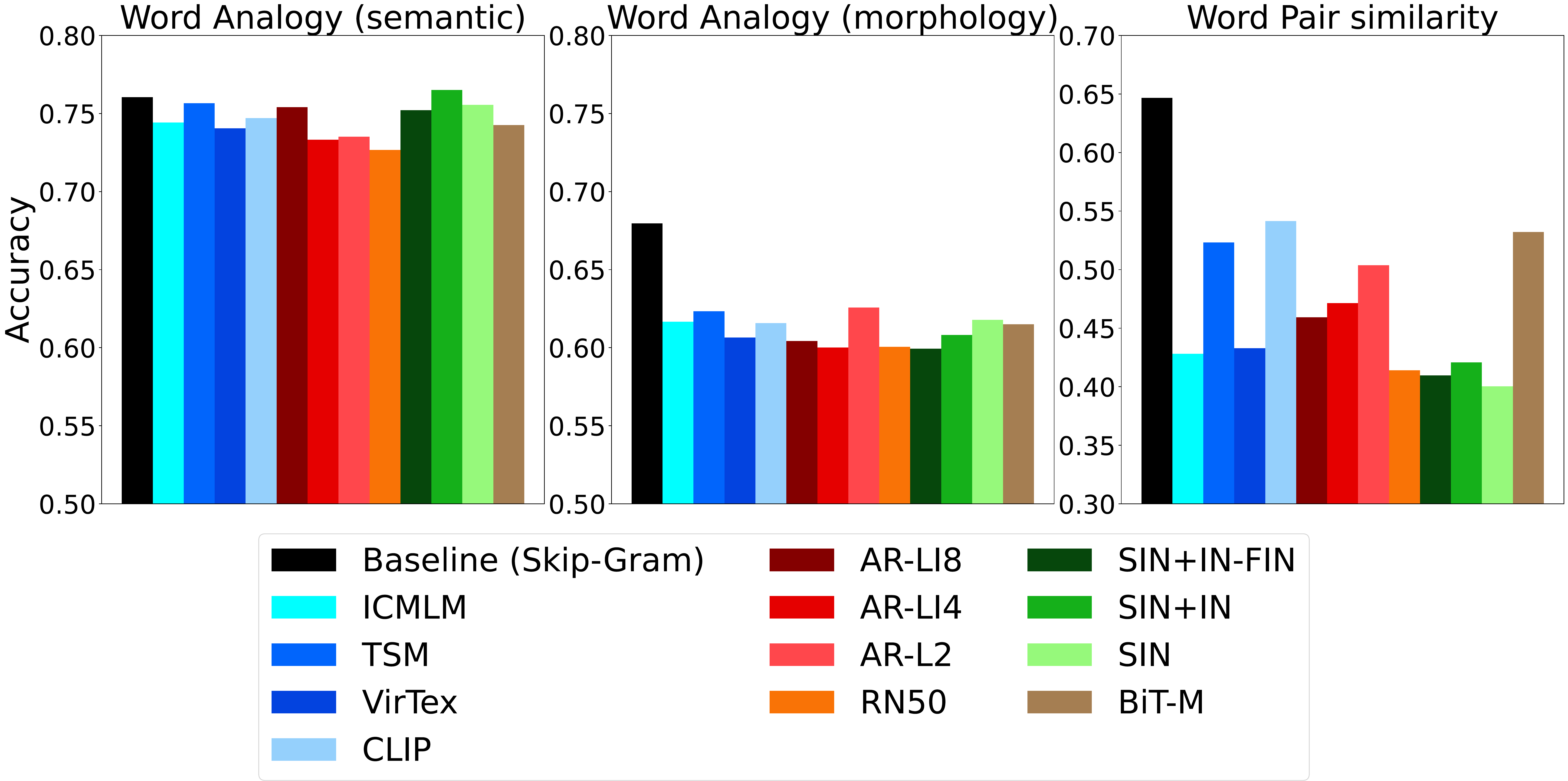}
    \caption{Semantic Word Analogy (such as ``son'', ``daughter'', ``boy'', ``girl''), Morphology Word Analogy (such as ``write'', ``writes'', ``work'', ``works'') and word pair similarity results for the visually constrained Skip-Grams.
    The Baseline is a vanilla Skip-Gram model (300 dimensions) where all 20,456 word embeddings are free to be learned.}
    \label{fig:linguistic-generalization}
\end{figure*}

We evaluate our Skip-Gram embeddings on two tasks: word analogies and word pair similarities.
\paragraph{Word Analogy}

This standard task \cite{Word2Vec} for evaluating the quality of word embeddings consists of quadruplets $\{A,B,C,D\}$ (e.g. ``man'', ``king'', ``woman'',``queen'') supporting the relation ``A is to B as C is to D''. The task consists in finding the 4th one given the first three, by solving the equation in the latent space:
$D = B-A+C$.
The more accurate the model, the better its representation. 
 We evaluate the word embeddings on the full dataset provided by \cite{Word2Vec} that we split in two different sets: \textit{morphology analogies} (such as ``write'', ``writes'', ``work'', ``works''), and \textit{semantic analogies} (such as ``son'', ``daughter'', ``boy'', ``girl''). If vision-language training helps ``ground'' the visually-derived word embeddings, we expect this grounding to be more helpful in the resolution of semantic, rather than morphology analogies.

\paragraph{Word Pair similarity}
Another task for evaluating the quality of word embeddings is to ask humans to rate the semantic similarity of pairs of words (e.g. on a scale of 0 to 10, how close is ``queen'' to ``king''? How close is ``queen'' to ``woman''? etc.) \cite{finkelstein2001placing} and then compute the same similarity evaluations in the latent spaces of the models. The higher the (Pearson) correlation between the pairwise similarities of a model and human pairwise similarity judgments, the better the representation of the model. 

\subsection{Results}

The baseline Skip-Gram produces the best word embeddings overall (black bars in Fig~\ref{fig:linguistic-generalization}). This is to be expected since the embeddings are learned freely, without any additional constraint during training.
Interestingly, this baseline advantage is weakest in the case of the semantic analogy task (Fig~\ref{fig:linguistic-generalization}, leftmost panel), where some of the vision and visio-linguistic models are on par with the baseline. This shows that the frozen vectors do not necessarily impede the performance when the analogies are defined semantically (and might thus be presumed to contain some visual component). However, even for these semantic analogies, vision or vision-language word embeddings never significantly surpassed the baseline performance.

In the word pair similarity task, networks show variable performance levels, but without a clear distinction between vision-only and vision-language models.
Among the visio-linguistic networks, CLIP and TSM, which are trained contrastively on a large amount of data (see Figure~\ref{fig:dataset-size-training}) have embeddings that correlate well with human word similarity judgements.
However, when compared with the vision-only models, we do not observe a clear-cut performance improvement. Indeed, the best vision-only model (BiT-M) is on par with CLIP and TSM.
Interestingly, by comparing the results from the Fig~\ref{fig:linguistic-generalization} rightmost panel to the data plotted in Fig~\ref{fig:dataset-size-training}, we observe that among our twelve models, the top six for the word pair similarity task (TSM, CLIP, BiT, and the three AR models) correspond to those models that were trained on the largest datasets.

For the analogy tasks (semantic and morphology), there is no particular trend. However in both cases, the best performing model (excluding the baseline) is a visual one: SIN+IN in the semantic case, and AR-L2 in the morphology case.


In summary, we find that multimodal training of visual features does not improve their usefulness for language tasks either, and we suggest that the amount of training data may be a more important factor for generalization.

\subsection{Legitimacy of the visual word embeddings}
In the previous results, for training the visually-guided word embedding models, we averaged the visual feature vectors over many examples for each class. This averaging can potentially alter the quality of the embeddings, e.g. by discarding important information about the feature distributions.
 Thus, we check the validity of these averaged feature vectors\footnote{We here test the 300d vectors after the PCA dimensionality reduction.}, by verifying that they remain useful in a vision context. We use these visual feature vectors as class prototypes and evaluate the corresponding nearest-neighbor classification accuracy on the ImageNet validation set\footnote{With the images regrouped into our 824 classes.} with a method similar to section \ref{section:few-shot}. For all models considered, classification accuracy was well above chance (p<0.01): this means that the class-specific vectors indeed remain useful as visual representations of their category.

 Furthermore, we computed the correlation between this visual classification accuracy of the word embedding, and the corresponding word analogy or word-pair similarity accuracy for each model. The resulting Pearson correlation coefficient was r=-0.0821 with the semantic Word Analogy performance, r=0.301 with the morphology Word Analogy performance, and r=0.797 with the Word Pair Similarity.

 The significant high correlation of visual classification with word-pair similarity performance might be caused by the visual component of the word similarity judgments performed by human subjects. Indeed, many ``similar words'' also entail similar visual features (tiger, jaguar, cat, feline), and so the word-pair similarity task may not be a pure language task.


\section{Discussion and Conclusion}

%
%
It is a highly appealing notion that semantic grounding could improve vision models, by introducing meaningful linguistic structure into their latent space, and thereby increasing their robustness and generalization properties.
Unfortunately, our experiments reveal that current vision-language training methods do not achieve this objective: the resulting multimodal networks are not better than vision-only models, neither for few-shot learning, transfer learning or unsupervised clustering, nor for adversarial robustness.
In addition, compared to vision-only models, the multimodal networks' visual representations do not appear to provide additional semantic information that could serve as a useful constraint for a word-embedding linguistic space.

The present inability of linguistic grounding methods to deliver their full promise does not imply that this cannot happen in the future. In fact, we believe that exploring the current models' performance and representations, as we do here, can help us understand their limitations and adjust our methods accordingly.
Specifically, we found that multimodal representations are neither visual nor linguistic, but are not really in-between either (Fig \ref{fig:tsne-dendrogram}). In CLIP and TSM, for instance, the contrastive learning objective encourages the visual and language streams to agree on a joint embedding of images and corresponding captions. However, such agreement, by itself, does not constrain either latent space to remain faithful to its initial domain. As a result, CLIP's (and TSM's) visual representations may discard information that could prove critical for transfer-learning to other visual tasks. If this is true, we predict that adding domain-specific terms to the multimodal loss function (e.g. self-supervision) could be a way to improve visual generalization, while retaining the advantages of multimodal training---possibly including semantic grounding.

\section*{Acknowledgements}
This research was supported by ANITI ANR grant ANR-19-PI3A-0004, AI-REPS ANR grant ANR-18-CE37-0007-01 and OSCI-DEEP ANR grant ANR-19-NEUC-0004.

\bibliography{output}
\bibliographystyle{acl_natbib}

\clearpage
\appendix

\section{Datasets}\label{appendix:datasets}
We briefly describe all the datasets used in our experiments. 

\subsection{CIFAR10 and CIFAR 100}
These datasets contain images of animals and objects comprising either 10 (CIFAR10) or 100 (CIFAR100) categories. All the 60,000 images --50,000 train and 10,000 test-- are of $32\times 32$ resolution with RGB color channels.

\subsection{CUB dataset}

Caltech-UCSD Birds, or CUB, dataset consists of 6033 images of 200 species of birds. Apart from the species name, the dataset also provides bounding boxes, approximate bird segmentation and attribute labels for each image, allowing for a finer analysis at a feature level. 

\subsection{MNIST}
Considered one of the simplest datasets, MNIST contains $28\times 28$ black and white images of handwritten digits from 0 to 9. It comprises of 60,000 training images and 10,000 test images.

\subsection{Fashion MNIST}
Based on article images from Zalando, an e-commerce platform, Fashion MNIST contains $28\times 28$ black and white images of 10 clothing categories. Designed with an aim to act as a ``direct drop-in replacement for the original MNIST'', it contains the same number of training and testing images as that of MNIST.

\subsection{StreetView House Numbers}
StreetView House Numbers, or SVHN dataset consists of images of digits from 0 to 9. Compared to MNIST, it is generally considered a more real-world dataset for optimizing neural networks since it contains images of digits in a more natural setting-- 600,000 colored images of digits provided by Google Street View images.

\end{document}